\documentclass[lettersize,journal]{IEEEtran}
\usepackage{amsmath,amsfonts}
\usepackage{algorithmic}
\usepackage{algorithm}
\usepackage{array}
\usepackage[caption=false,font=normalsize,labelfont=sf,textfont=sf]{subfig}
\usepackage{textcomp}
\usepackage{stfloats}
\usepackage{url}
\usepackage{verbatim}
\usepackage{graphicx}
\usepackage{cite}
\usepackage{xcolor}
\usepackage{bm}
\usepackage{booktabs}
\usepackage{multirow}
\usepackage{makecell}
\begin{document}

\title{A Transferability Metric Using Scene Similarity and Local Map Observation for DRL Navigation}

\author{Shiwei Lian and Feitian Zhang
\thanks{Shiwei Lian is with the Department of Advanced Manufacturing and Robotics,                College of Engineering, Peking University, Beijing, 100871, China,
        {\tt\small lianshiwei@stu.pku.edu.cn}}%
\thanks{Feitian Zhang is with the Department of Advanced Manufacturing and Robotics, and the State Key Laboratory of Turbulence and Complex Systems, College of Engineering, Peking University, Beijing, 100871, China,
        {\tt\small feitian@pku.edu.cn}}%
}



\maketitle

\begin{abstract}
While deep reinforcement learning (DRL) has attracted a rapidly growing interest in solving the problem of navigation without global maps, DRL typically leads to a mediocre navigation performance in practice due to the gap between the training scene and the actual test scene. To quantify the transferability of a DRL agent between the training and test scenes, this paper proposes a new transferability metric --- the scene similarity calculated using an improved image template matching algorithm. Specifically, two transferability performance indicators are designed including the global scene similarity that evaluates the overall robustness of a DRL algorithm and the local scene similarity that serves as a safety measure when a DRL agent is deployed without a global map. In addition, this paper proposes the use of a local map that fuses 2D LiDAR data with spatial information of both the agent and the destination as the DRL observation, aiming to improve the transferability of DRL navigation algorithms. With a wheeled robot as the case study platform, both simulation and real-world experiments are conducted in a total of 26 different scenes. The experimental results affirm the robustness of the local map observation design and demonstrate the strong correlation between the scene similarity metric and the success rate of DRL navigation algorithms.
\end{abstract}


\section{Introduction}
Autonomous navigation is an indispensable ability for mobile robots to find a collision-free path towards destinations. To enhance navigational performances in unknown environments, deep reinforcement learning (DRL) has emerged as a promising approach for its generalization ability \cite{b13, ral4,b11, b12}. Particularly, DRL has demonstrated great potential in navigation without a global map that takes multi-modal sensory data as observations \cite{b16, b17, b18, b20}.
With low data sample efficiency, most DRL navigation algorithms first train policies in simulators and then deploy these trained policies to robots in the real world. However, the gap between simulation and reality or the difference between the training scene and the test scene often leads to poor navigation performance when a well-trained DRL is transferred to or deployed in a new environment \cite{kenzo, IPAPRec}. Specifically, in autonomous navigation, the DRL algorithm trains a reactive strategy that takes the action with the highest cumulative reward in training. The trained strategy, statistically speaking, has a lower navigation success rate when transferred to the test environment due to the differences between the training and test scenes in, for example, the number and the size of obstacles \cite{survey2}. 

While a number of DRL navigation methods have been investigated to increase the success rate when applied in new environments, limited research has looked into the quantification of the transferability of DRL navigation algorithms. This paper proposes a transferability metric for the DRL navigation algorithms by quantifying the similarity between the training scene and the test scene.
Specifically, this paper proposes two scene similarity performance indicators using the improved image template matching algorithm to quantify the transferability of the DRL navigation algorithm. 
The global scene similarity, calculated from the global maps of the training and test scenes, is designed to evaluate the overall transferability or robustness of different navigation algorithms. The local scene similarity, taking the collected local obstacle maps in the test scenes, serves as a safety indicator when a trained agent is deployed in a new environment without a global map. 
In addition, this paper designs a robust DRL navigation algorithm using the local map as the observation that fuses 2D LiDAR data, the agent position, and the destination position together onto the local map as the input to the action network. Experimental results through diverse scenes both in simulation and the real world further show the effectiveness of the proposed scene similarity metric and local map-based DRL navigation in quantifying and enhancing algorithmic transferability, respectively.

The main contributions of the paper are threefold.
First, a novel scene similarity metric involving global and local measures is proposed to quantify the transferability of DRL navigation using the improved image template matching algorithm. To the best of the authors' knowledge, this is the \textbf{first} effort to design such a transferability metric using scene similarity for DRL navigation.
Second, different from common designs using range measurements as observation inputs, a robust DRL navigation algorithm is proposed using the fused local map as the input, which allows interchanging the LiDAR sensor with different fields of views (FoVs) and angular resolutions, thus exhibiting higher transferability at deployment time.
Third, extensive experiments with diverse scenes constructed in both simulation and the real world are conducted. The experimental results support the validation of the proposed transferability metric and show the robustness of the designed DRL algorithm using the local map as the observation.

\section{Related Work}

\subsection{Robot Navigation}

This paper focuses on navigation tasks in unknown environments without a global map. Given only the local perception of the robot, traditional path/motion planning algorithms such as artificial potential fields (APF)\cite{apf} and dynamic window approach\cite{dwa} would likely fail due to the local optimum. In addition, their navigation performance is highly dependent on their selected parameters, while it is generally difficult to update the parameters for an unknown environment\cite{b11}.

To overcome the limitations of the traditional methods, DRL algorithms have been applied to enhance navigational performances. Sensor data used in the DRL navigation of mobile robots includes but not limited to 2D LiDAR \cite{b11, b12, b16, b17, IPAPRec, Tmech2, iTD3}, RGB images \cite{Tmech1}, and depth images \cite{b13, ral4, b20}. 2D LiDAR data is most commonly used due to the minor sim-to-real gap and extensively represented in the form of vectors \cite{b12, b16, b17, IPAPRec}. This representation results in a fixed sensory FoV and a fixed-size measurement as the observation. 

To relax the constraint on 2D LiDAR observations and enhance the DRL navigation performance, researchers have investigated a number of alternative observation designs. For example, Leiva \textit{et al.} \cite{b11} converted LiDAR data into 2D point clouds with variable sizes as observations and achieved improved navigational performances. Pfeifer \textit{et al.} \cite{ocmap} and Yao \textit{et al.} \cite{b14} converted 2D LiDAR data to occupancy maps to address more complex and dynamic environments showing great potential in robot navigation with sensor fusion.

This paper proposes a local map-based DRL navigation algorithm independent of a global map. The 2D LiDAR data, the agent position, and the destination position are all fused onto the local map as the observation. Convolutional neural networks (CNNs) are utilized to efficiently extract the features of the fused spatial data and learn a robust navigation policy. The local map-based navigation algorithm is independent of the use of the transferability metric.

\subsection{DRL Transferability}

To enhance the transferability of DRL, many studies have been conducted to bridge the gap between the training and test performances, including but not limited to domain randomization\cite{dr1, dr2, dr3, chaffre}, learning from demonstration\cite{b21}, domain adaptation\cite{da1, da2, Bharadhwaj}, and meta-RL\cite{b22}.

For instance, Chaffre \textit{et al.}\cite{chaffre} designed a DRL policy with incremental environment complexity to reduce the need for additional training in the real world. Hieu \textit{et al.}\cite{b21} developed transfer learning with demonstrations to accelerate the training process when an autonomous vehicle entered a new environment. Bharadhwaj \textit{et al.}\cite{Bharadhwaj} performed adversarial domain adaptation to learn an encoder that generated the same distribution of latent states over real images as the simulated images and then fine-tuned the learnt policy in real environments. Luo \textit{et al.}\cite{b22} proposed meta-learning on the latent space algorithm for rapid adaptation of DRL-based visual navigation to new observations. The studies above mainly focused on the design of DRL algorithms to increase the success rate when transferred from a training scene to a test scene, but did not look into the quantification or the measure of the DRL transferability.

Chebotar \textit{et al.}\cite{dr1}, on the other hand, quantified the DRL transferability by introducing a discrepancy function evaluated by computing the difference between the simulated and real-world robotic arm trajectories using weighted $l_1$ and $l_2$ norms. This discrepancy measure calculates the difference given the motion state inputs, however, cannot be directly applied to calculate the difference of two navigation scenes.

This paper focuses on finding an appropriate metric to quantify or measure the transferability of DRL navigation algorithms but without looking into how to re-train or update the algorithms in the new environment. This paper proposes a new transferability metric for robot navigation using the improved image template matching algorithm to measure the similarity between the training scenes and the real test scenes. Two straightforward and easy-to-implement performance indicators are designed including the global and local scene similarities. The global scene similarity is used to evaluate the transferability of different DRL navigation algorithms. The local scene similarity indicates the safety of the local map-based DRL navigation algorithm when deployed in a new environment without a global map.

\section{Problem Description}

This paper applies the DRL method to design a navigation algorithm for a mobile robot without the prior knowledge of a global map and provides a measure of the transferability of the designed algorithm when applied to new environments. We assume that at time instant $t$, the robot agent takes the sensor measurements and its relative position to the destination as the observation $\bm{o}_t$. Apply the DRL algorithm to learn a policy $\pi$ with weights $\theta$ in a training scene $S_{\rm train}$. When applied to a test scene $S_{\rm test}$, the policy is expected to map the observation to a suitable agent action $\bm{a}_t$ that steers the agent towards the destination without collision, \textit{i.e.},

\begin{equation}
\bm{a}_t = \pi_{\theta | S_{\rm train}} (\bm{o}_t \mid S_{\rm test})
\label{eq_policy}
\end{equation}

The gap between the training and test scenes is typically the major cause of failure of a learnt policy when deployed in new environments. To quantify the transferability of the DRL policy, scene similarity between the test scene and the training scene is measured. Given the global map of the test scene ${M}_{{\rm test}}$ and the global map of the training scene ${M}_{{\rm train}}$, a scoring function $f_1()$ calculates the global scene similarity score $SS_{\rm global}$ through

\begin{equation}
SS_{\rm global} = f_1({M}_{{\rm train}}, {M}_{{\rm test}})
\label{eq_ss_global}
\end{equation}

\noindent This global metric is used to evaluate the performance of different DRL navigation algorithms when deployed in different test scenes with different $SS_{\rm global}$. 

When the global map of the test scene is not available, which is the typical case in practice, the agent only obtains local information through sensor measurements. A scoring function $f_2()$ calculates the local scene similarity score $SS_{\rm local}$ from the collected observations $\bm{o}_{\rm test}$ in the test scene and the global map of the training scene ${M}_{{\rm train}}$, \textit{i.e.},

\begin{equation}
SS_{\rm local} = f_2({M}_{{\rm train}},\bm{o}_{\rm test})
\label{eq_ss_local}
\end{equation}

\noindent This local metric is expected to predict the success rate of the proposed local map-based DRL navigation algorithm when deployed in new environments.

To improve the transferability performance of DRL navigation, the problem to be addressed is summarized as to design the global and local scene similarities (Eqs.~(\ref{eq_ss_global}) --(\ref{eq_ss_local})) to quantify the navigation transferability as well as to incorporate the DRL policy with local map-based observations (Eq.~(\ref{eq_policy})) for enhanced robustness with respect to different test scenes.

\section{Local Map-Based DRL Navigation}

\subsection{Deep Reinforcement Learning}

Motion planning is considered as a Markov decision process (MDP). The MDP is represented by the tuple ${(\mathcal{S, A, T,} r, \gamma)}$, where ${\mathcal{S} = \{s\}}$ is the state space, ${\mathcal{A} = \{a\}}$ is the action space, $\mathcal{T}$ is the state transition function indicating the probability distribution of the next state when performing action $a$ in the current state $s$, $r$ is the reward function, and ${\gamma \in [0, 1]}$ is the discount factor. At each discrete time step, the agent selects and takes an action based on the current state. The agent then receives a reward and transitions to the next state. The objective of the agent is to learn an optimal policy ${\pi^*(a\mid s)}$ that maximizes the expected return. 

In this paper, we use DQN\cite{b6} and TD3 \cite{b7} as example DRL algorithms to learn navigation with discrete and continuous actions, respectively. Both algorithms are sample efficient with the replay buffer to collect experience during training and have been proved with high navigation performance \cite{b13,iTD3}. The implementation of DRL navigation is specified as follows. The observation or the input of DRL is the local map $\bm M$, which will be discussed in detail in Section \ref{section_local_map}.

\textbf{Action space}: A differential drive robot is selected as the testing platform in this study. The action space of the differential drive robot includes a translational velocity $v_t$ and a rotational velocity $\omega_t$, \textit{i.e.} ${\{ \bm{a_t}:= [v_t, \omega_t] \}}$. Specifically, for DQN, three discrete actions are considered, \textit{i.e.}, ${v=0.15 \text{ m/s}}$, ${\omega \in \{-1.0, 0.0, 1.0\} \text{ rad/s}}$. For TD3, continuous actions are considered, \textit{i.e.}, ${v \in [0., 0.25]\text{ m/s}}$, ${\omega \in [-1.0, 1.0] \text{ rad/s}}$.

\textbf{Reward design}: The objective of the agent is to use the oncoming local map observations to reach the destination point as fast as possible without colliding with obstacles. The reward at time $t$, denoted by $r^t$, is designed as

\begin{equation}
r^t = r_{\rm step}^t + \left \{
\begin{array}{rcl}
r_{\rm succ}^t  &      &  \text{if } d_t < 0.25 \rm m\\
r_{\rm nav}^t     &      & \text{if } d_{t} > d_{t-1}\\
r_{\rm coll}^t     &      & \text{if collision}
\end{array} \right.
\label{eq_reward}
\end{equation}

\noindent Here, $d_t$ is the distance from the agent to the goal at time $t$. $r_{\rm step}^t$ gives a negative reward at each time step so that the agent reaches the goal destination as quickly as possible. $r_{\rm succ}^t$ generates a positive reward when the agent enters a neighborhood of the goal to encourage success. $r_{\rm nav}^t$ is used to encourage approaching the goal and the agent receives a negative reward when it moves away from the goal. $r_{\rm coll}^t$ is used to avoid the collision of the agent with obstacles. It generates a negative reward when a collision occurs. In this paper, ${r_{\rm step}^t =0}$, ${r_{\rm succ}^t=1}$, ${r_{\rm nav}^t=-0.001}$ and ${r_{\rm coll}^t=-1}$ is set for DQN. ${r_{\rm step}^t =-0.05}$, ${r_{\rm succ}^t=5}$, ${r_{\rm nav}^t=-0.1}$ and ${r_{\rm coll}^t=-5}$ is set for TD3. $r_{\mathrm{step}}^t$ for DQN is set to 0 for the reason that a negative $r_{\mathrm{step}}^t$ increases the training time.

\textbf{Termination conditions}: During training, an episode will be terminated if one of the following conditions is satisfied. 1) The agent reaches the goal. 2) The agent collides with an obstacle. 3) The number of the time steps exceeds the maximum time step $\rm T_{max}$. For DQN, $\rm T_{max}$ is set to 10,000. For TD3, $\rm T_{max}$ is set to 2,000.

\subsection{Using Local Maps as Observations}\label{section_local_map}
The 2D LiDAR data, the agent position and the goal position are all spatial information. They are all fused onto the local map as the observation or the input of the DRL algorithm. Using the local map as the observation allows for learning a more robust DRL navigation policy and interchanging LiDAR sensors with different angular resolutions and FoVs at the deployment time.

The local map is denoted by $\bm{M} = \left[M_{\rm o}, M_{\rm p}, M_{\rm g}\right] \in \mathbb{R}^{3 \times H \times W}$. Here, $H$ and $W$ represent the height and width of the map, respectively. $M_{\rm o}$ is the obstacle map converted from 2D LiDAR data. $M_{\rm p}$ is the position map, indicating the agent's current position. $M_{\rm g}$ is the goal map, indicating the distance and the orientational angle of the destination point from the agent. $M_{\rm o}$, $M_{\rm p}$ and $M_{\rm g}$ are merged together to fuse all the data and generate the local map.

\begin{figure}[tpb]
\centerline{\includegraphics[width=0.9\linewidth]{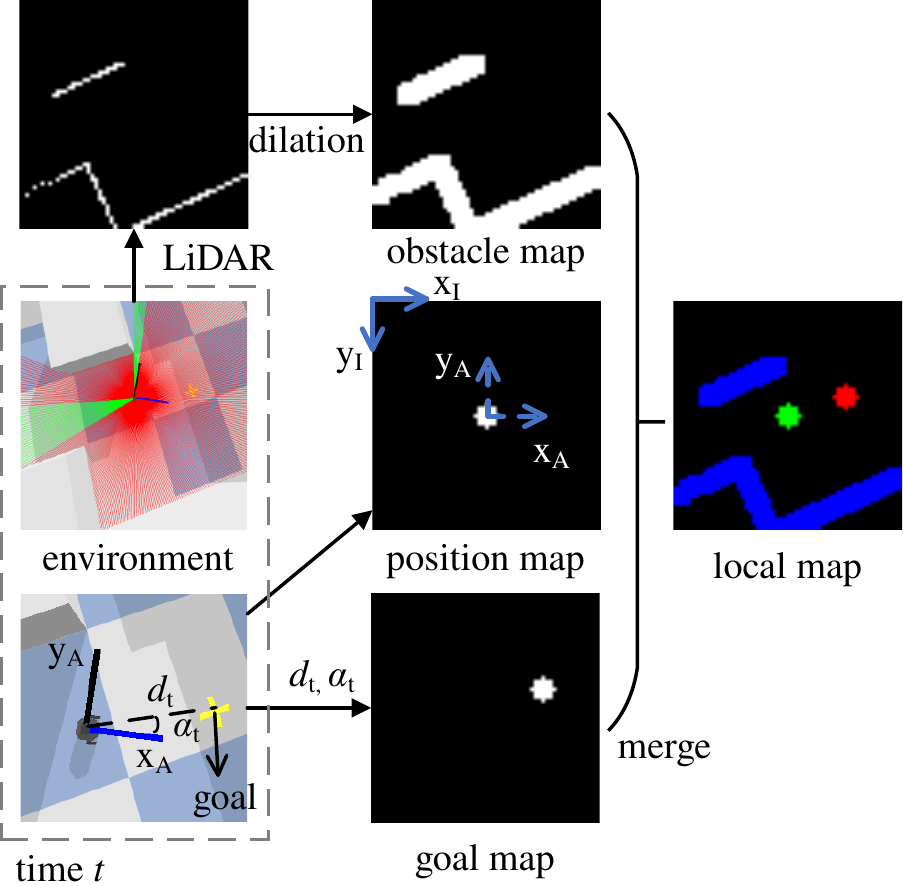}}
\caption{The schematic of the generation of the local map at time $t$.}
\label{local_map}
\end{figure}

The process of generating the local map is illustrated in Fig.~\ref{local_map}. The obstacle map $M_{\rm o}$ is converted from the LiDAR data. The LiDAR data $\{ (\rho_t, \theta_t)\}$ received at time $t$ typically consists of the distance $\rho_t$ and the orientation angle $\theta_t$ of an obstacle with respect to the agent expressed in the polar coordinates. The LiDAR polar coordinates $(\rho_{\rm A},\theta_{\rm A})$ is convertible to the image coordinates $(x_{\rm I},y_{\rm I})$ by

\begin{equation}
\left[
\begin{array}{rcl}
x_{\rm I}  \\
y_{\rm I}
\end{array} \right] = 
\left[
\begin{array}{rcl}
\rho_{\rm A} \cos{(\theta_{\rm A})} /R + W/2 \\
- \rho_{\rm A} \sin{(\theta_{\rm A})} /R + H/2
\end{array} \right]
\label{eq_coord}
\end{equation}

\noindent Here, $R$ (meter/pixel) is the resolution of the map, $H$ and $W$ represent the height and width of the map, respectively, and the subscripts I and A indicate the image coordinate and the agent coordinate, respectively. The obstacle map is generated by a dilation operation \cite{dilation} ignoring the LiDAR data outside. The dilation operation increases the thickness of LiDAR data, repairs small breaks in the map, and enlarges the features of small obstacles. Compared to the classic input of LiDAR data as a fixed-size vector, the obstacle map is independent of the sensor resolution and FoV, thus allowing the LiDAR data with arbitrary measurement sizes. The local map observation is adaptable to other sensor modules for sensor fusion. 

The position map $M_{\rm p}$ encodes the agent's shape and current position. In this paper, a circle with a radius of the same pixel as the agent is used to represent the agent's shape. The circle in $M_{\rm p}$ is set at the center of the local map at any time step helping the network recognize the danger of collisions with nearby obstacles and the navigational orientation towards the goal destination.

The goal map $M_{\rm g}$ uses the circle to represent the locations of the destination. There are cases where the target point is outside the boundary of the map $M_{\rm g}$. To ensure $M_{\rm g}$ is non-empty, the distance of the goal with respect to the agent at time $t$, denoted by $d_t$, is clipped to $d_t'$ by

\begin{equation}
d_t' = \min\left( d_t, \frac{RW}{2|\cos \alpha_t|}, \frac{RH}{2|\sin \alpha_t|}\right)
\label{eq_dt}
\end{equation}

\noindent Here, $R$ (meter/pixel) is the resolution of the map, $H$ and $W$ represent the height and width of the map, respectively, and $\alpha_t$ is the orientation angle of the goal with respect to the agent at time $t$. The coordinates of the center of the circle $(x_{\rm Ig}, y_{\rm Ig})$ in $M_{\rm g}$ are obtained by converting $d_t'$ and $\alpha_t$ following the same coordinate transformation described in Eq.~(\ref{eq_coord}).

\begin{figure}[tpb]
\centerline{\includegraphics[width=0.9\linewidth]{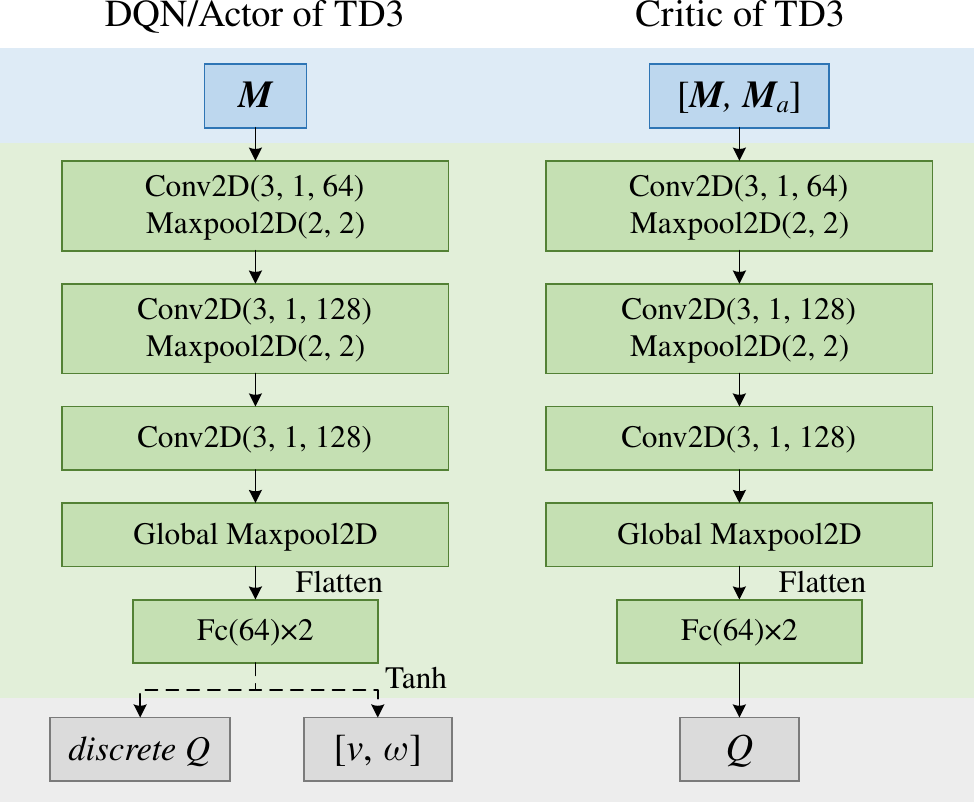}}
\caption{The schematic of the adopted network architecture for DQN and the actor-critic network of TD3. Conv2D (kernel size, stride, number of kernels) is the 2D convolutional layer. Maxpool2D (kernel size, stride) is the 2D max pooling layer. Fc (number of hidden units) is the fully connected layer. The activation function is ReLU.}
\label{fig_network}
\end{figure}

The local map fuses all information to take advantage of the powerful feature extraction capability of CNNs. The network architectures of DQN and the actor-critic network of TD3 are shown in Fig. \ref{fig_network}. The three convolutional layers of DQN and the TD3 actor network efficiently extract the features of the input local map $\bm{M}$. The feature map is then processed by the global max pooling layer to extract global features. The global features are flattened and fed into the fully connected layers. The output of DQN is the $Q$ value of the three discrete actions. The output of the TD3 actor network is the predicted continuous actions $[v, \omega]$ using the activation function of Tanh. 

The critic network of TD3 requires additional input, \textit{i.e.}, the agent's action. The input of the critic is either formulated as a numerical vector that concatenates the agent's action and the extracted feature vector of CNNs output $[v, \omega]$ or a fused local map that combines the local map with the action map designed as ${\bm{M}_a = [M_{\rm v}, M_{\rm \omega}]}$ by replacing non-zero values of $M_{\rm p}$ with action values. We find that fusing $\bm{M}_a $ with the local map $\bm{M}$ as the input of the TD3 critic network achieves better navigation performance. The output of the critic network is the state-action estimate value.

\section{Scene Similarity Metric Based On Improved Image Template Matching}


\subsection{Image Template Matching Algorithm}\label{matching_algo}

The image template matching is an image processing method to search and find the maximum match between a template image and another larger image. Sliding the template image over the image of interest, the algorithm calculates the similarity between the sub-images and the template image and outputs the position with the maximum similarity as the matching result. The similarity is calculated as the normalized correlation coefficient $R_{I, T}(x,y)$ by

\begin{equation}
\begin{split}
&R_{I, T}(x,y)= \\
&\frac{ \sum\limits_{x_T} \sum\limits_{y_T} (T'(x_T,y_T) \cdot I'(x+x_T,y+y_T)) }{\sqrt{\sum\limits_{x_T} \sum\limits_{y_T} T'(x_T,y_T)^2 \cdot \sum\limits_{x_T} \sum\limits_{y_T} I'(x+x_T,y+y_T)^2} }
\end{split}
\label{eq10}
\end{equation}

\noindent where

\begin{equation}
 T'(x_T,y_T)=T(x_T,y_T) - \frac{\sum\limits_{i=0}^{w-1} \sum\limits_{j=0}^{h-1} T(i,j)}{w \cdot h} 
\label{eq8}
\end{equation}

\begin{equation}
\begin{split}
I'(x&+x_T,y+y_T)= \\
&I(x+x_T,y+y_T) - \frac{\sum\limits_{i=0}^{w-1} \sum\limits_{j=0}^{h-1} I(x+i,y+j)}{w \cdot h}
\end{split}
\label{eq9}
\end{equation}

\noindent Here, $I$ and $T$ represent the image to be matched and the template image, respectively. ${R_{I, T}(x, y)}$ is the similarity of the template $T$ to the sub-image of the image $I$ at ${(x,y)}$. ${T(x_T, y_T)}$ represents the pixel value at ${(x_T, y_T)}$ of the image~$T$. $w$ and $h$ represent the width and height of the image~$T$, respectively. The best match score denoted by ${R^{*}(I, T)}$ is defined by

\begin{equation}
R^*(I, T) = \max_{x, y} R_{I, T}(x,y)
\label{eq11}
\end{equation}

The template matching above is suitable for the case when the template and the target object in the image have the same orientation. Matching often fails when there is an orientational difference between the target and the template. Therefore, the template image needs to be rotated to generate a series of templates with different rotation angles $\phi$. The best match score is then updated by

\begin{equation}
R^*\left(I, T\right) = \max_{x, y, \phi} R_{I, T_\phi}\left(x,y\right)
\label{eq12}
\end{equation}

\subsection{Global Scene Similarity}
Currently, most DRL navigation algorithms achieve good navigation performance when the test scene is sufficiently similar to the training scene. However, some abnormal behaviors of DRL agents are often observed when tested in dissimilar scenes \cite{IPAPRec}. We propose a global scene similarity performance indicator based on the image template matching algorithm described in Section~\ref{matching_algo}. This indicator quantifies the similarity between the training scene and the test scene to evaluate the transferability of DRL navigation algorithms in both similar and dissimilar scenes. 

\begin{figure}[tpb]
\centerline{\includegraphics[width=0.95\linewidth]{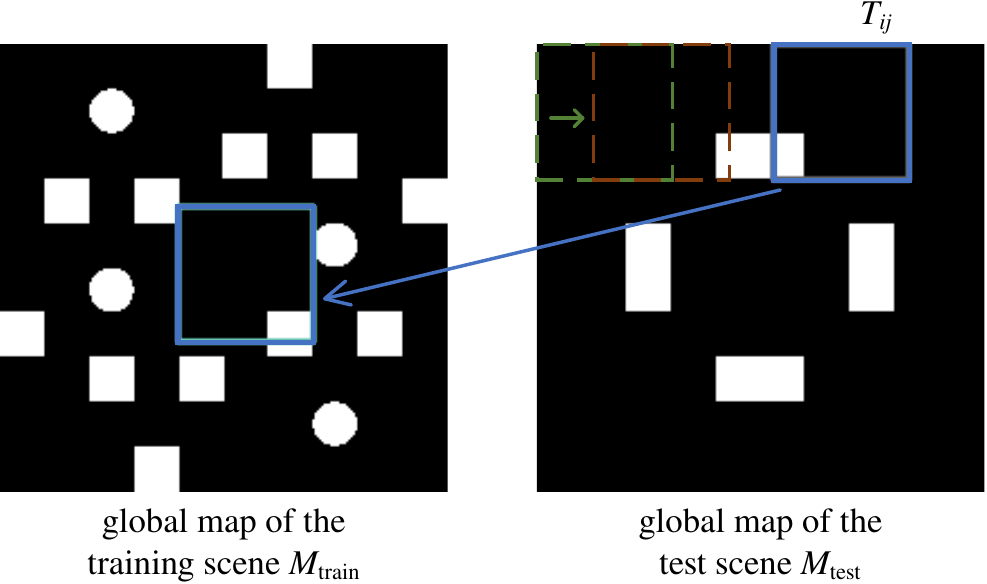}}
\caption{Illustration of the calculation of the global scene similarity.}
\label{global_match}
\end{figure}

The calculation of the global scene similarity metric is illustrated in Fig.~\ref{global_match}. The image to be matched is the global map of the training scene denoted by ${M}_{{\rm train}}$. The template is the sub-images $T_{ij}$ of the global map of the test scene ${M}_{{\rm test}}$ generated through a sliding window. The global scene similarity score $SS_{\rm global}$ between ${M}_{{\rm train}}$ and ${M}_{{\rm test}}$ is calculated as the averaged best match scores by 

\begin{equation}
SS_{\rm global} = \frac{1}{N_1 N_2}\sum_{i}^{N_1} \sum_{j}^{N_2} R^*(M_{\rm train}, T_{ij})
\label{eq_ss_global2}
\end{equation}

\noindent where
\begin{equation}
T_{\rm ij} =
\begin{bmatrix}
M_{\rm test}(is_{x}, js_{y}) & \cdots  & M_{\rm test}(is_{x}+w-1,  \\
&  &  js_{y})\\
\vdots & \ddots   & \vdots \\
M_{\rm test}(is_{x},  &  \cdots & M_{\rm test}(is_{x}+w-1,   \\
js_{y}+h-1)  &   &  js_{y}+h-1)   \\
\end{bmatrix}
\label{eq_ss_t}
\end{equation}

\noindent Here, $R^*(M_{\rm train}, T_{ij})$ is the best match score between ${M}_{{\rm train}}$ and $T_{ij}$ using Eq.~(\ref{eq12}) with the rotation angle ${\phi \in \{10^{\circ}, 20^{\circ}, ..., 360^{\circ}\}}$. $T_{ij}$ is the sub image of $M_{\rm test}$ generated through a sliding window with strides $s_x$ and $s_y$ to reduce computation. ${M_{\rm test}(is_{x}, js_{y})}$ represents the pixel value at ${(is_{x}, js_{y})}$ of ${M_{\rm test}}$. $w$ and $h$ represent the width and height of the image $T_{ij}$, respectively.

\subsection{Local Scene Similarity}

The local scene similarity is proposed to quantify the transferability of the local map-based navigation algorithm when deployed in a new environment without the global map. Without the global map of the test scene, only the local obstacle maps collected along the agent's trajectory are available. The metric calculates the similarity between the actual observations in the test scene and the observations in the training scene. Since the global map of the training scene is available and easily obtained, the proposed local scene similarity utilizes the global obstacle map for the convenience of calculation instead of the local map. 

The calculation of the local scene similarity metric is based on the image template matching method described in Section~\ref{matching_algo}. The image to be matched is the global obstacle map of the training scene denoted by $M_{\rm o}^{\text{global}}$. The template is the local obstacle maps converted from the LiDAR measurements in the test scene. The image template matching method finds the averaged best match scores between the local obstacle maps collected in the test scene and the global obstacle map of the training scene.



\begin{figure}[tpb]
\centerline{\includegraphics[width=0.85\linewidth]{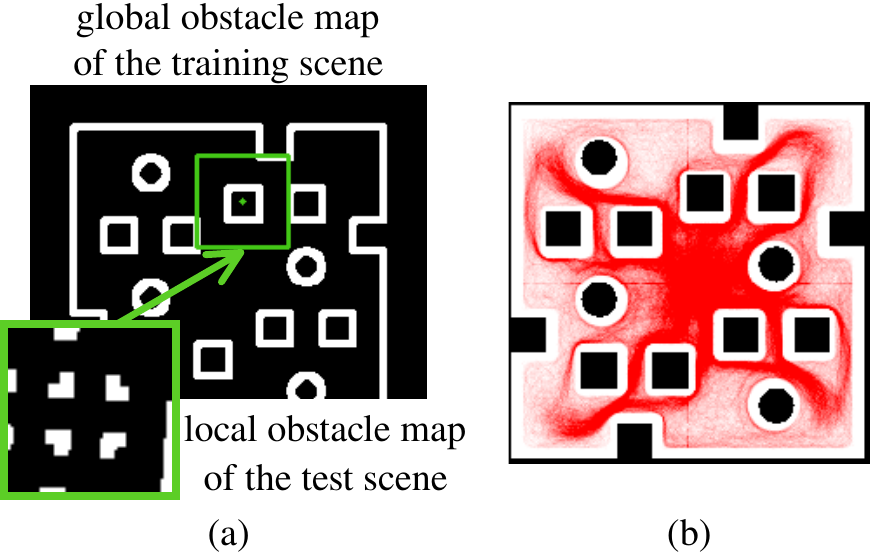}}
\caption{(a) Illustration of problem \#1 using conventional image template matching method. The center of the best match lies within the obstacle region that the robot agent cannot reach in practice. (b) Illustration of problem \#2 using conventional image template matching method. It shows all trajectories stored in the replay buffer. A point in darker red indicates that the observation at that point is more likely to be sampled in the training and thereby better learnt than a point in lighter red.}
\label{problem}
\end{figure}

We find that the direct use of template matching causes the following two problems.
First, as shown in Fig.~\ref{problem}(a), the center of the best match sub-image sometimes lies within the obstacle region in the training scene, and the agent cannot reach such a location during the training process.
Second, Fig.~\ref{problem}(b) shows all the trajectories of the agent stored in the replay buffer during training with the darkness in red indicating the total number of arrivals. We observe a significant variance among different locations regarding how many times an observation is used in the training. Therefore, it is unfair that different positions in the global obstacle map carry the same level of importance.

To solve the aforementioned problems, the template matching method is updated by introducing a weight matrix in the calculation of the match score. The weight is calculated based on how many times the agent reaches a specific position on the global map during training, which indicates the importance of the match score at the corresponding position. The weight at position $(x,y)$ in $M_{\rm o}^{\text{global}}$ is calculated by

\begin{equation}
\text{Wt}(x, y) =
\text{clip}\left(
\frac
{\text{clip}\left(\#(\text{arrivals})(x, y), 0, N_{\text{max}}\right)}
{N_{\text{max}}}, 0.5, 1 \right)
\label{eq13}
\end{equation}

\noindent where ${\#(\text{arrivals})(x, y)}$ is the total number of times the agent reaches the position $(x,y)$ in $M_{\rm o}^{\text{global}}$ during the training. To normalize the weight and prevent it from being too small, two layers of clip functions are used. In the first layer, $N_{\text{max}}$ is the total number of the agent reaching a position that we consider as the threshold to give the highest importance. In the second layer, [0.5,1] is used to define the range of the weight values. The introduction of the weight matrix helps the metric to distinguish specific training processes.

The scene similarity score $SS$ is formulated as follows. First, the agent collects $N$ local obstacle maps ${\{M_{{\rm o}i}, i \in [0, N) \}}$ from the test scene. Then, the averaged best match score between  ${\{M_{{\rm o}i}, i \in [0, N) \}}$ and the global obstacle map ${M_{\rm o}^{\text{global}}}$ of the training scene is calculated by

\begin{equation}
SS = \frac{1}{N}\sum_{i}^N \max_{x, y, \phi}\text{Wt}\left(x+\frac{W}{2},y+\frac{H}{2}\right) R_{M_{\rm o}^{\text{global}}, M_{{\rm o}i\phi}}\left(x,y\right)
\label{eq14}
\end{equation}

\noindent where $SS$ is the scene similarity score, $W$ and $H$ are the width and height of the local obstacle map, respectively, and the rotation angle ${\phi \in \{1^{\circ}, 2^{\circ}, ...,360^{\circ}\}}$. 

Considering that the similarity score of the training scene itself {$SS_{\rm train}$} varies, to debias the influence of the training scene, the local scene similarity $SS_{\rm local}$ is therefore defined as the relative scene similarity score or the difference between the similarity score calculated in the test scene $SS_{\rm test}$ and the similarity score calculated in the training scene {$SS_{\rm train}$}, \textit{i.e.},

\begin{equation}
SS_{\rm local} = SS_{\rm test}-SS_{\rm train} 
\label{eq15}
\end{equation}

\section{Case Study on Local Scene Similarity Metric}
In this section, the effectiveness of the local scene similarity metric in quantifying the transferability of the local map-based DQN navigation algorithm is shown through both simulation and real-world experiments in diverse test scenes when no global map of the test scenes is available. 

\subsection{Case Study Implementation Details}\label{sec_im}
The configuration of the local map and the hyper parameters of DQN and TD3 are listed in Table \ref{tab1}. DQN and TD3 are implemented in PyTorch and trained using the Adam optimizer. The generation of the local maps and the calculation of the scene similarity metric use the OpenCV library.

\begin{figure*}[tpb]
\centerline{\includegraphics[width=0.95\linewidth]{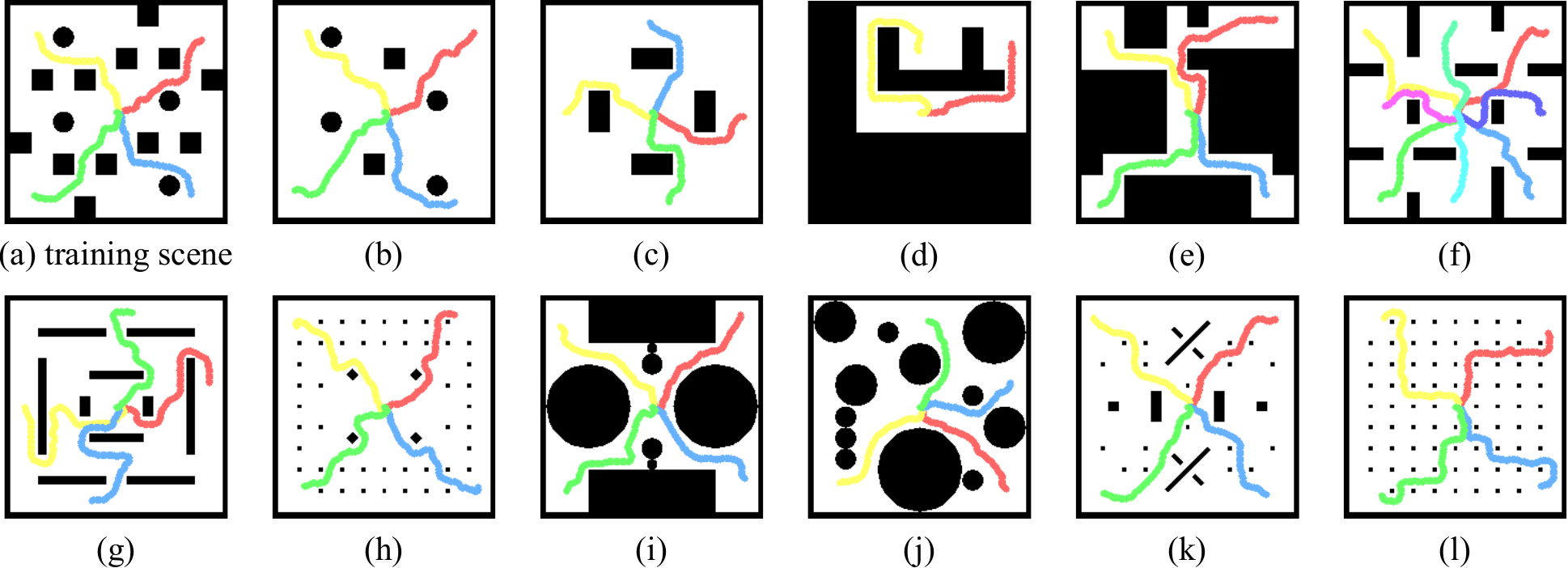}}
\caption{12 test scenes constructed in the PyBullet simulator and the sampled trajectories of the robot agent trained by the local map-based DQN algorithm. All the test scenes have the dimension of 10m$\times$10m. The order of the scenes is sorted from the highest to the lowest $SS_{\rm local}$.}
\label{scenarios}
\end{figure*}

The training and the testing in simulation are carried out in the PyBullet simulator on a computer with an i7-12700F CPU and an NVIDIA GeForce RTX 3070 GPU. The agent is a differential drive robot equipped with a laser range finder that has a max range of 3.5m, a $360^{\circ}$ FoV and 360 measurement readings. A total of 12 scenes with a 10m$\times$10m two-dimensional space are constructed in simulation as shown in Fig.~\ref{scenarios}. The training of DQN and TD3 takes approximately 18 hours and 3 hours for each agent in scene (a), respectively.

\begin{table}[tbp]
\caption{Hyper Parameters}
\begin{center}
\begin{tabular}{lll}
\toprule
& \textbf{Parameter} & \textbf{Value} \\
\hline
\multirow{3}{*}{local map}&size ($H, W$) & (60, 60) \\
&resolution $R$ & 0.05 m/pixel \\
&kernel size of dilation & 5\\
\hline
\multirow{7}{*}{DQN}&learning rate & $10^{-4}$ \\
&discount factor $\gamma$ & 0.99 \\
&initial exploration & 1.0 \\
&final exploration & 0.05 \\
&batch size & 64 \\
&replay buffer size & $10^6$ \\
&number of training episodes & 4500 \\
\hline
\multirow{10}{*}{TD3}&actor, critic learning rate & $5\times10^{-5}$, $10^{-3}$ \\
&discount factor $\gamma$ & 0.99 \\
&initial exploration noise & 1.0 \\
&final exploration noise & 0.05 \\
&policy exploration noise & 0.2 \\
&policy update delay & 2 \\
&soft update factor $\tau$ & 0.005 \\
&batch size & 64 \\
&replay buffer size & $2\times10^5$ \\
&number of training episodes & 2500 \\
\bottomrule
\end{tabular}
\label{tab1}
\end{center}
\end{table}

\subsection{Simulation Results}

The simulation environment is set up as described in Section \ref{sec_im}. 17 different agents are first trained using DQN in the same training scene (Fig.~\ref{scenarios}(a)) with the same hyper parameters listed in Table \ref{tab1} but different DQN initial weights and exploration processes. 
Following that, the agents are deployed in 12 test scenes that include the training scene itself as shown in Fig.~\ref{scenarios}. The comparison results between the calculated local scene similarity scores $SS_{\rm local}$ and the navigation success rates are analyzed and presented to support the validation of the effectiveness of the proposed scene similarity metric in quantifying the transferability of the navigation algorithm.


\begin{figure}[tpb]
\centerline{\includegraphics[width=1.0\linewidth]{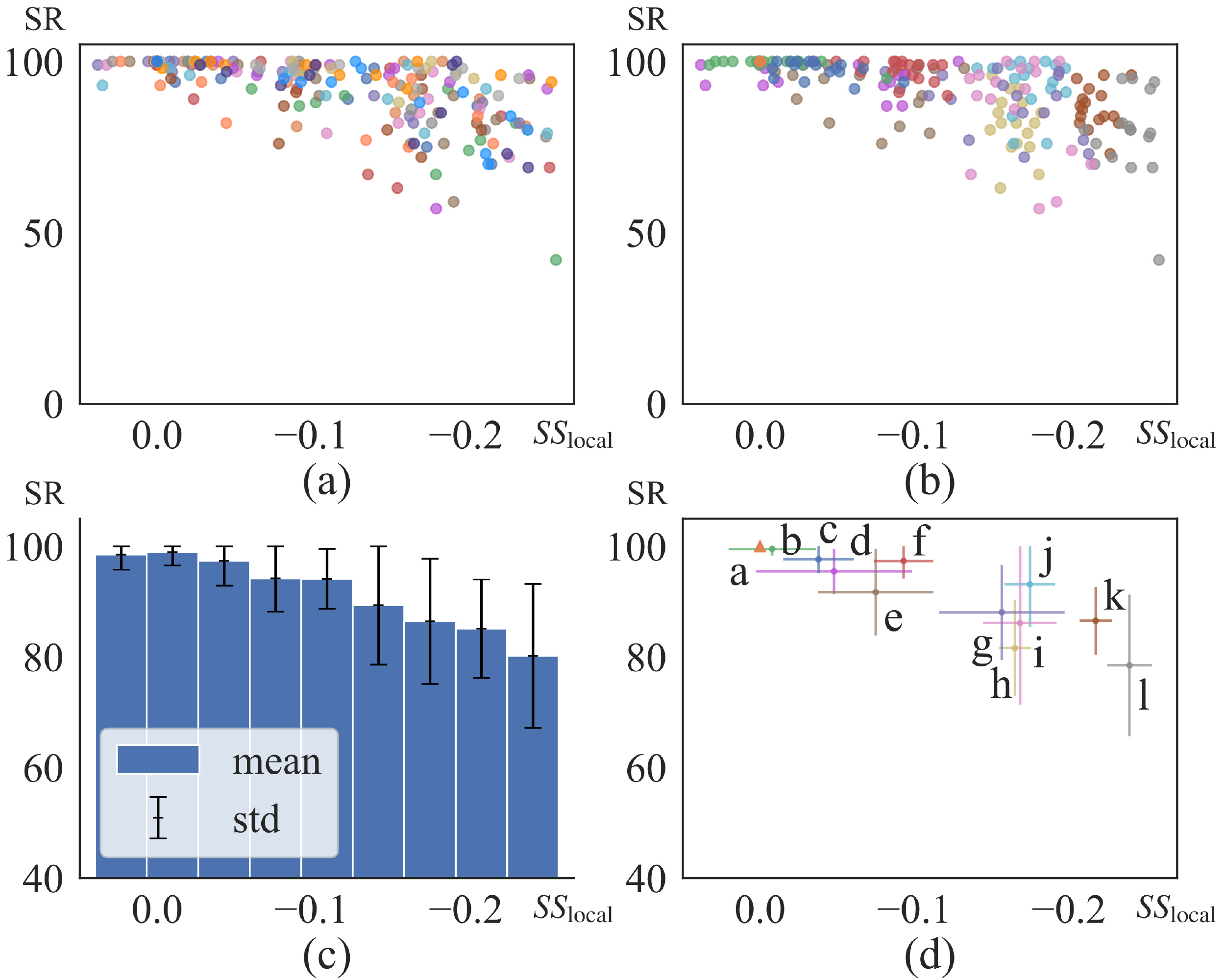}}
\caption{The statistical results of the relationship in simulation between the scene similarity metric $SS_{\rm local}$ and the navigation success rate (SR) in test scenes. (a) $SS_{\rm local}$ vs SR for the 17 trained agents represented by 17 different colors. (b) $SS_{\rm local}$ vs SR for the 10 test scenes represented by 10 different colors. (c) The mean and standard deviation of SR over all the test data expressed in 9 equally distributed intervals of $SS_{\rm local}$ ranging from 0.04 to -0.26. (d) The mean and standard deviation of SR and $SS_{\rm local}$ over all test scenes. The labels of the scenes are the same as in Fig.~\ref{scenarios}.}
\label{simulaiton}
\end{figure}

Simulation results show that the local map-based DQN navigation algorithm successfully steers the robot to the destination in the 12 test scenes with various success rates. In each scene, 100 destination points are randomly generated to calculate the success rate.

Figure~\ref{simulaiton}(a) shows the scatter plot between the navigation success rate and the local scene similarity metric over all 17 agents represented by 17 different colors. We observe that with a decreased local scene similarity score $SS_{\rm local}$, the navigation success rate in general has a wider distribution or a larger variation along with a decreased mean value. Figure~\ref{simulaiton}(c) classifies all the test data in Fig.~\ref{simulaiton}(a) according to $SS_{\rm local}$ into nine equal intervals ranging from 0.04 to -0.26, showing the mean and standard deviation of the success rate for each interval. We observe a similar trend that an increased uncertainty in the success rate is correlated with a decreased scene similarity. 

Figure~\ref{simulaiton}(b) presents the simulation results according to the scene, using 12 different colors to represent the 12 test scenes. Figure~\ref{simulaiton}(d) shows the mean and the standard deviation of the success rate and $SS_{\rm local}$ for each scene. The scene alphabet labels are consistent with that in Fig.~\ref{scenarios}. Generally speaking, as the local scene similarity score decreases, the mean of the success rate shows a decreasing trend, and the variance shows an increasing trend. 

It is noteworthy that scene~j has a lower similarity than scene~e, but the navigation success rate is approximately equal. We conjecture that the navigation success rate is not only related to the similarity between the training and test scenes but also to the complexity of the scene itself. Although scene~e has a high similarity, it contains narrower passages, and thus the agent is more likely to collide during navigation. In contrast, scene~j is less similar to the training scene, but with more open space, thus making the navigation task easier.

\subsection{Real-World Experiment}

\begin{figure*}[thpb]
\centerline{\includegraphics[width=1.0\linewidth]{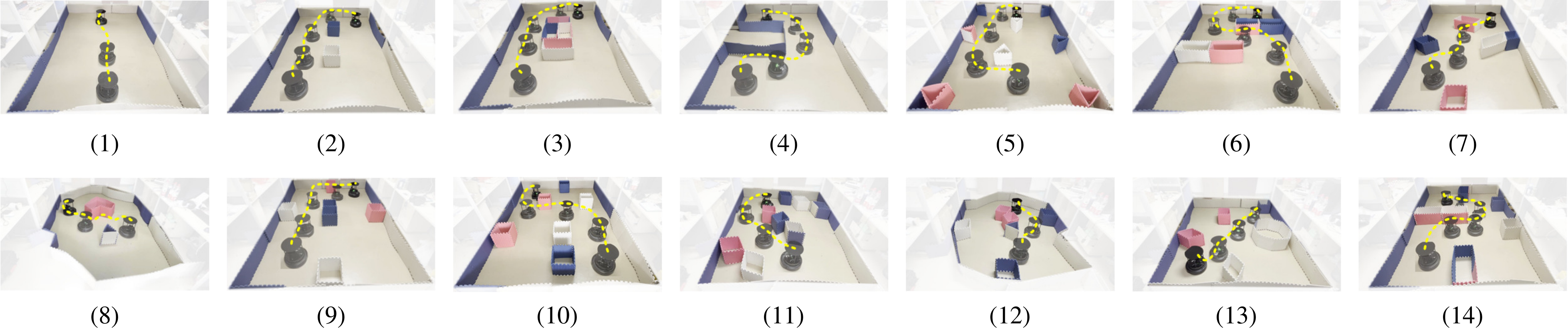}}
\caption{The navigation trajectories in different real-world test scenes. The order of the scenes is sorted from the highest to the lowest similarity, and the labels of the scenes are the same as in Fig.~\ref{real_world}.}
\label{realworld_senario}
\end{figure*}

\begin{figure}[tpb]
\centerline{\includegraphics[width=1.0\linewidth]{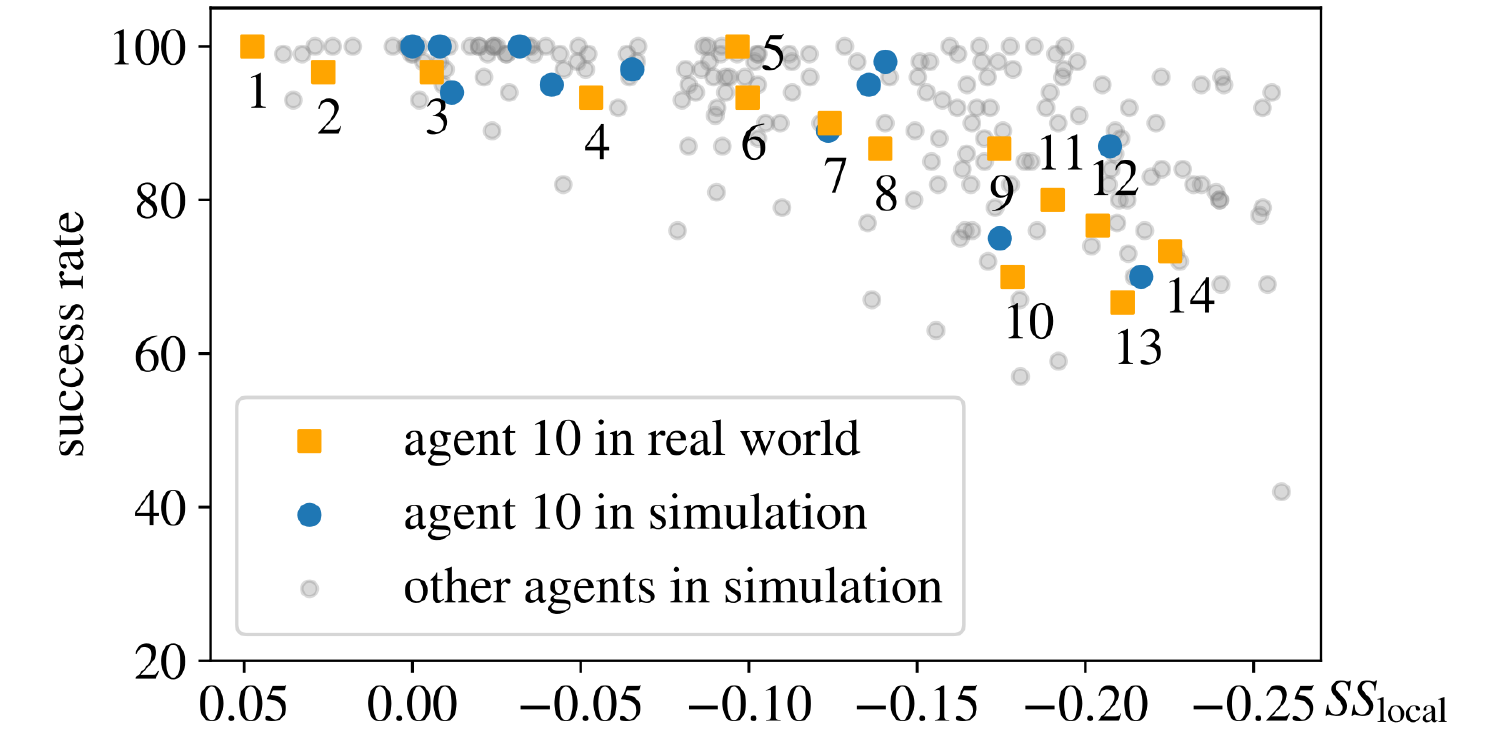}}
\caption{The local scene similarity scores and the navigation success rate in the real-world test scenes. The labels in the figure are consistent with those in Fig.~\ref{realworld_senario}.}
\label{real_world}
\end{figure}

A TurtleBot4 wheeled robot with a maximum linear velocity of 0.31 m/s and a maximum angular velocity of 1.90 rad/s is used to experimentally test the proposed navigation algorithm and the local scene similarity metric. It is equipped with a RPLIDAR A1M8 LiDAR featuring a maximum measuring range of 12 m, a FoV of $360^\circ$ and an angular resolution $\le 1^\circ$. The navigation algorithm runs on a laptop that communicates with the robot in real time. The laptop has an i7-12700H CPU and an NVIDIA GeForce RTX 3060 GPU.

A total of 14 test scenes are constructed in the lab. The scenes and the sampled navigation trajectories are shown in Fig.~\ref{realworld_senario}. The order of the scenes is sorted by the local similarity score, arranged from the highest to the lowest. Without loss of generality, the trained agent \#10 from the simulation is selected and used in the experiment. In each scene, 30 goal destinations are generated to perform the proposed local map-based DQN navigation algorithm. 

Figure~\ref{real_world} shows the scatter plot of the experimental results between the relative scene similarity score and the navigation success rate. The proposed local map-based DQN navigation algorithm successfully steers the robot toward its destinations in the real world. The success rate varies with the local similarity score following the same pattern as observed in the simulation. As the local scene similarity score decreases, the success rate has a smaller mean value and a larger standard deviation, indicating increased uncertainties in the model.

The simulation and experimental results confirm that the local scene similarity metric and the navigation success rate of the proposed local map-based navigation algorithm are strongly correlated. The proposed scene similarity metric has a great potential in providing guidance for the design of training scenes and the transfer learning algorithm so that an improved navigation performance (e.g., increased safety) can be achieved.

\section{Case Study on Global Scene Similarity and Local Map Observation}
In this section, the local map-based DRL navigation is tested and compared to other DRL algorithms with classic and state-of-the-art observations of vector-based LiDAR measurements in simulation using the global scene similarity $SS_{\rm global}$. In addition, ablation experiments are conducted to show the benefit of fusing all spatial data onto the local map. The configurations of the LiDAR sensor are changed to show that the local map is independent of the FoV and the angular resolution of the sensor at the deployment time. Finally, a comparison experiment with traditional algorithms is conducted.

\subsection{Comparisons With Other Observations}\label{com_vs_other}
To showcase the transferability and robustness of using the local map as the observation, comparison studies are conducted with selected classic and state-of-the-art observations. The global scene similarity $SS_{\rm global}$ is adopted as the evaluation metric. 

\begin{figure}[tpb]
\centerline{\includegraphics[width=1.0\linewidth]{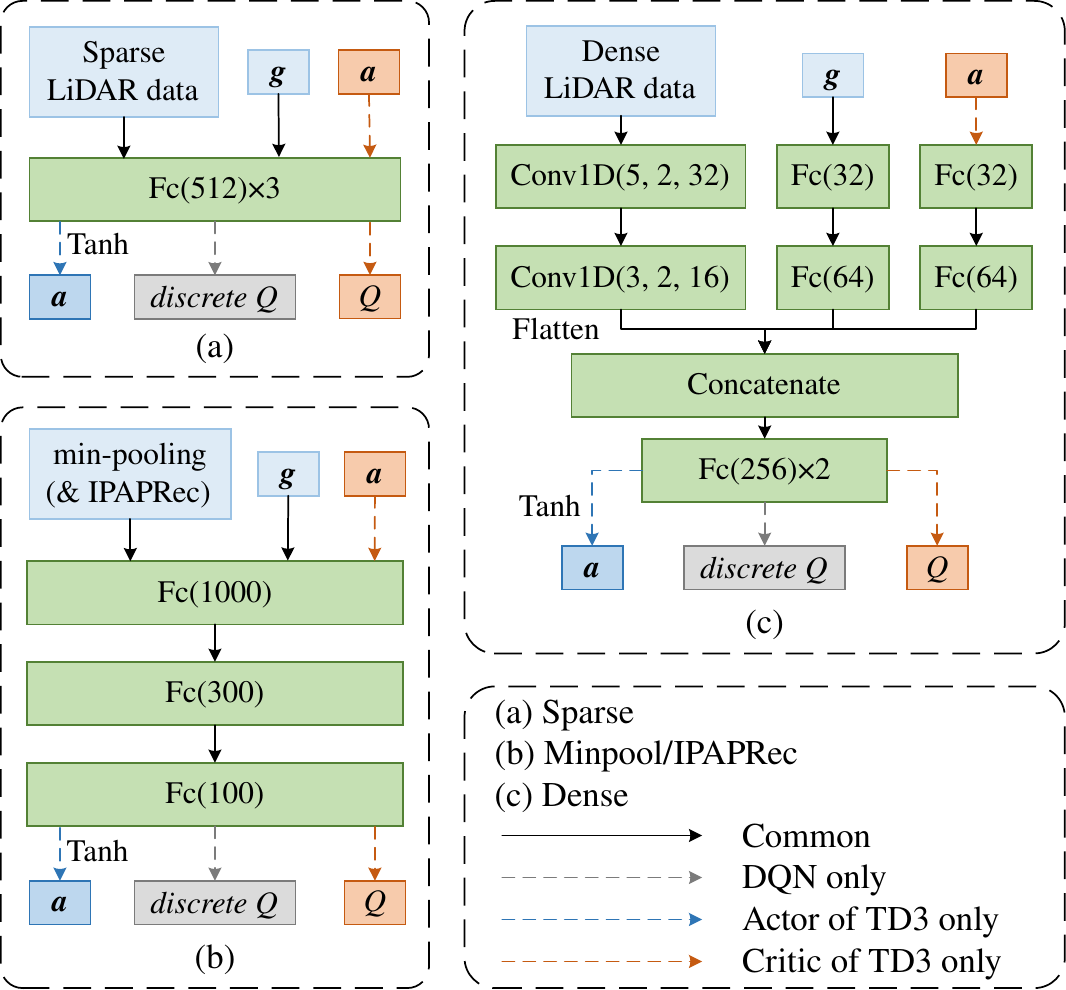}}
\caption{The different observations used to compare and the schematic of adopted network architectures of DQN and TD3. \textit{\textbf{g}} represents the goal information. $\bm{a}$ represents the action. Conv1D (kernel size, stride, number of kernels) is the 1D convolutional layer. Fc (number of hidden units) is the fully connected layer. The activation function is ReLU.}
\label{fig_network3}
\end{figure}

The selected observations and corresponding network architectures are shown in Fig.~\ref{fig_network3}.
The first observation shown in Fig.~\ref{fig_network3}(a) is the sparse LiDAR data proposed in \cite{b16}. The network architecture similar to \cite{b16} is used to train. The spare LiDAR data is a normalized 10-dimensional vector uniformly sampled from the raw laser measurements from angle $-90^\circ$ to $90^\circ$. The vector is concatenated with the goal information ${\bm{g}=[d, \alpha]}$ as the input, where $d$ and $\alpha$ are the distance and the orientation angle of the goal with respect to the agent, respectively.
The second observation shown in Fig.~\ref{fig_network3}(b) is the LiDAR data after the min pooling operation proposed in \cite{b17}. The similar architecture proposed in \cite{b17} is used to train. The operation compresses the raw 360 laser measurements into a 36-dimensional vector. The third observation is the LiDAR data after the min pooling and the IPAPRec operation proposed in \cite{IPAPRec}, which is a powerful input preprocessing approach with adaptively parametric reciprocal functions. The initial value of IPAPRec trainable parameter $\beta_i$ is set to -1 in this paper.
The last observation shown in Fig.~\ref{fig_network3}(c) is the most commonly used dense LiDAR data, which uses the raw laser range measurements as a 360-dimensional vector. The adopted network architecture is similar to \cite{b11} which utilizes 1D convolutional layers to extract features of laser measurements.

\begin{figure}[tpb]
\centerline{\includegraphics[width=1.0\linewidth]{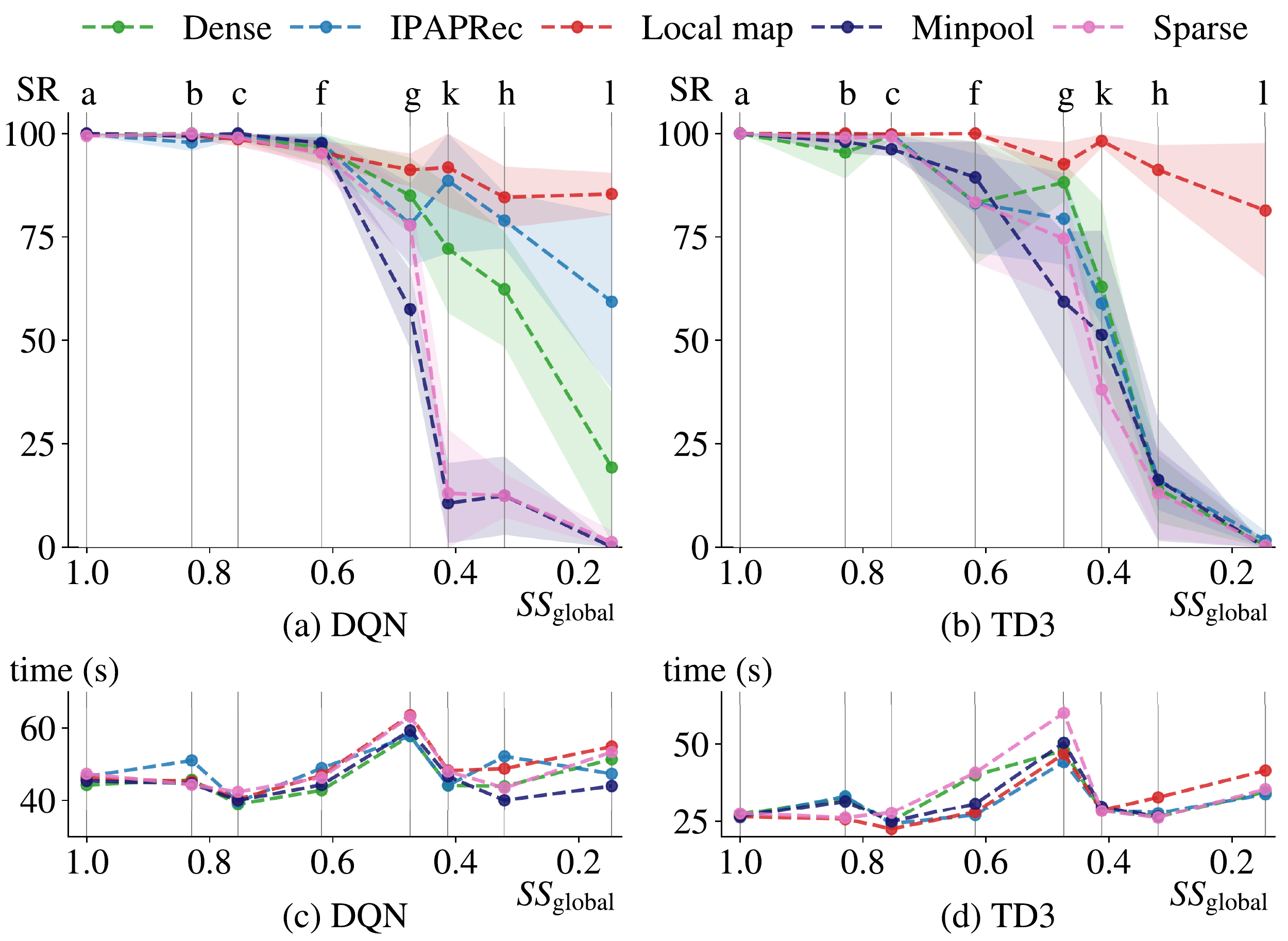}}
\caption{The average navigation success rates (SR) and the average navigation time using selected DRL observations in test scenes with respect to $SS_{\rm global}$. (a) SR of DQN. (b) SR of TD3. (c) The navigation time of DQN. (d) The navigation time of TD3.  The labels of the scenes are the same as in Fig.~\ref{scenarios}.}
\label{fig_comp}
\end{figure}

For each observation, 5 independent agents are trained using DQN and TD3 with the hyper parameters listed in Table \ref{tab1} and the reward function in Eq.~(\ref{eq_reward}). The trained agents are deployed in 8 test scenes in Fig.~\ref{scenarios}. Those test scenes are selected for clear presentation of the results considering that their global scene similarity scores $SS_{\rm global}$ with respect to the training scene differ considerably apart from each other. In each test scene, the same 100 destination points are generated for each agent. The average success rate and the average navigation time of each DRL observation with respect to $SS_{\rm global}$ are shown in Fig~\ref{fig_comp}. 

As shown in Fig.~\ref{fig_comp}(a) and Fig.~\ref{fig_comp}(b), the success rates of all DRL navigation algorithms using different observations show a decreasing trend as $SS_{\rm global}$ decreases. However, compared to other observations, using the local map achieves significantly better navigation performance with much higher navigation success rates in test scenes with lower $SS_{\rm global}$. Although there is no small obstacle in the training scene (scene~a), the local map-based DRL navigation still avoids colliding with small obstacles when deployed in test scenes h, k and l. The results confirm the enhanced robustness and transferability of the local map observation design for DRL navigation. The reason for the higher transferability of the local map is that it converts LiDAR data and the agent's position to the map, which helps the agent recognize the shapes of the surrounding obstacles as well as the colliding situation by recognizing the overlap between the obstacles and current position on the map. In comparison, Minpool, for example, only takes the minimum LiDAR measurement within a certain angle interval, not able to reflect the shape of obstacles in the angle interval and resulting in collisions when the shapes of the obstacles are dissimilar or more complex than the training scene. In Fig.~\ref{fig_comp}(c) and Fig.~\ref{fig_comp}(d), we observe that the navigation time of the local map in scenes (h) and (l) is slightly longer than other observations. We conjecture the reason is that the local map-based navigation takes safer and more conservative actions to avoid collisions in scenes with lower similarity scores.

\begin{figure}[tpb]
\centerline{\includegraphics[width=1.0\linewidth]{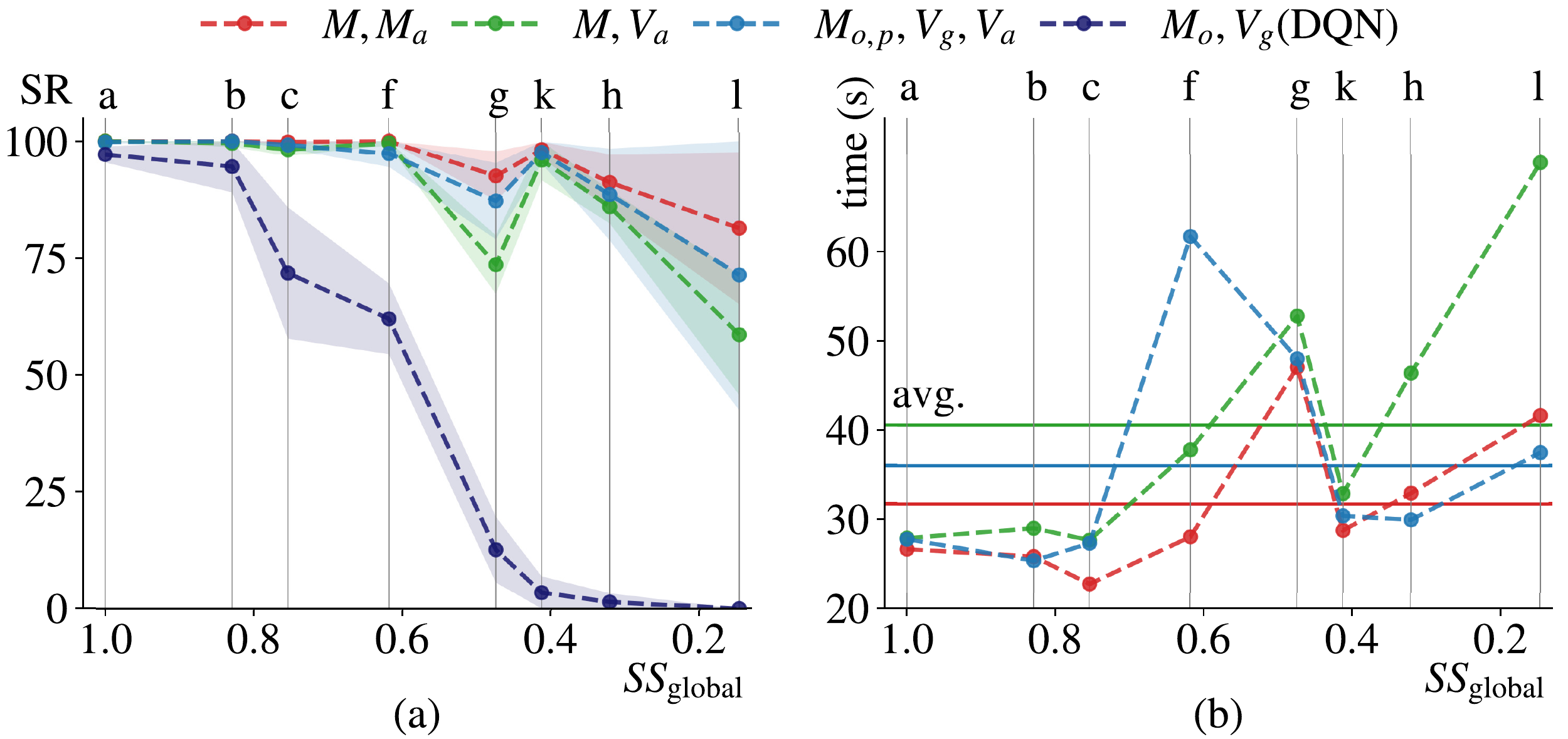}}
\caption{The average navigation success rates (SR) and the average navigation time using the local map input and other vector-type inputs of the goal and/or velocity data in test scenes with respect to $SS_{\rm global}$. The labels of the scenes are the same as in Fig.~\ref{scenarios}.}
\label{fig_ablation}
\end{figure}

\subsection{Ablation Study}
While the goal and velocity information are often modeled as a vector input, the local map fused all information onto a map, which helps the CNNs learn all the spatial information together. The proposed local map $\bm{M}, \bm{M}_{a}$ is compared to other inputs that use the goal and/or velocity as a vector and the results are shown in Fig.~\ref{fig_ablation}. We use $M$ to denote the map input and $V$ to denote the vector input. All the inputs are trained using the TD3 algorithm in Fig.~\ref{fig_network} except $M_{\rm o}, \bm{V}_{\rm g}$ being trained using DQN because the TD3 agent fails to learn a good policy when using this kind of input. The four selected inputs are detailed as follows.

1) $\bm{M}, \bm{M}_{a}$ denotes the use of the proposed local map $\bm{M}$. It converts the velocity information to the action map $\bm{M}_{a}$ as the input to the TD3 critic network. 
2) $\bm{M}, \bm{V}_{a}$ inputs the action as a vector $\bm{V}_{a}$ to the TD3 critic network and concatenates it with the extracted features of the local map $\bm{M}$ after CNNs. 
3) $M_{\rm o, p}, \bm{V}_{\rm g}, \bm{V}_{a}$ adds the robot position onto the obstacle map \cite{b13} denoted by $M_{\rm o, p}$, and inputs goal and velocity information as vectors $\bm{V}_{\rm g}$ and $\bm{V}_{a}$, respectively. 
4) $M_{\rm o}, \bm{V}_{\rm g}$ only inputs the obstacle map $M_{\rm o}$ and the goal information as a vector $\bm{V}_{\rm g}$.

As shown in Fig.~\ref{fig_ablation}, the proposed local map input $\bm{M}, \bm{M}_{a}$ achieves the best overall performance with a shorter average navigation time and higher average navigation success rates in almost all test scenes. $M_{\rm o}, \bm{V}_{\rm g}$ performs worst since it only inputs the obstacle map without the agent's position, thus very challenging for the network to learn the colliding situations with obstacles.

\subsection{Changing Sensor FoV and Angular Resolution}
Converting the LiDAR data to the local map allows an arbitrary number of LiDAR measurements. The FoV and the angular resolution of the LiDAR sensor are changed at the deployment time to test if the local map-based DRL navigation algorithm still performs robustly well with respect to different sensor setups. The average navigation success rates and average navigation time using different sensor setups tested in scene (a) of Fig.~\ref{scenarios} are listed in Table \ref{tab2}. The local map-based DRL navigation algorithm maintains a high success rate in almost all presented LiDAR setups. Only when the number of measurements is significantly small (\# of Meas.=30), the navigation success rate of the DQN-based algorithm decreases slightly. The results show the robustness and transferability of the local map-based navigation algorithm that allows the interchange of the sensor's FoV and angular resolution at the deployment time.

\begin{table}[tbp]
\caption{The mean and standard deviation of navigation success rate and navigation time in scene (a) with respect to LiDAR sensor configurations, \textit{i.e.}, the FoV and the number of laser measurements (\# of Meas.).}
\begin{center}
\begin{tabular}{cc|cc|cc}
\toprule
\multicolumn{2}{c|}{Variant} & \multicolumn{2}{c|}{DQN} & \multicolumn{2}{c}{TD3}\\
FoV & \makecell[c]{ \# of \\ Meas.} &  SR (\%) & Time (s) &  SR (\%) & Time (s)\\
\hline
360 & 360 & 99.8$\pm$0.4 & 46.4$\pm$1.8 & 100.0$\pm$0.0 & 26.7$\pm$0.8 \\
360 & 260 & 100.0$\pm$0.0 & 46.8$\pm$2.3 & 100.0$\pm$0.0 & 26.7$\pm$0.8 \\
360 & 160 & 99.6$\pm$0.8 & 46.9$\pm$1.8 & 100.0$\pm$0.0 & 26.7$\pm$0.8 \\
360 & 60 & 97.0$\pm$2.7 & 46.6$\pm$2.6 & 100.0$\pm$0.0 & 26.7$\pm$0.9 \\
360 & 30 & 95.4$\pm$1.9 & 45.8$\pm$2.1 & 99.4$\pm$1.2 & 27.1$\pm$1.0 \\
270 & 270 & 99.8$\pm$0.4 & 46.4$\pm$1.8 & 100.0$\pm$0.0 & 26.7$\pm$0.8 \\
180 & 180 & 99.8$\pm$0.4 & 46.5$\pm$1.3 & 99.2$\pm$1.2 & 27.0$\pm$0.7 \\
\bottomrule
\end{tabular}
\label{tab2}
\end{center}
\end{table}

\subsection{Comparison With Traditional Local Planners}

\begin{figure}[tpb]
\centerline{\includegraphics[width=0.8\linewidth]{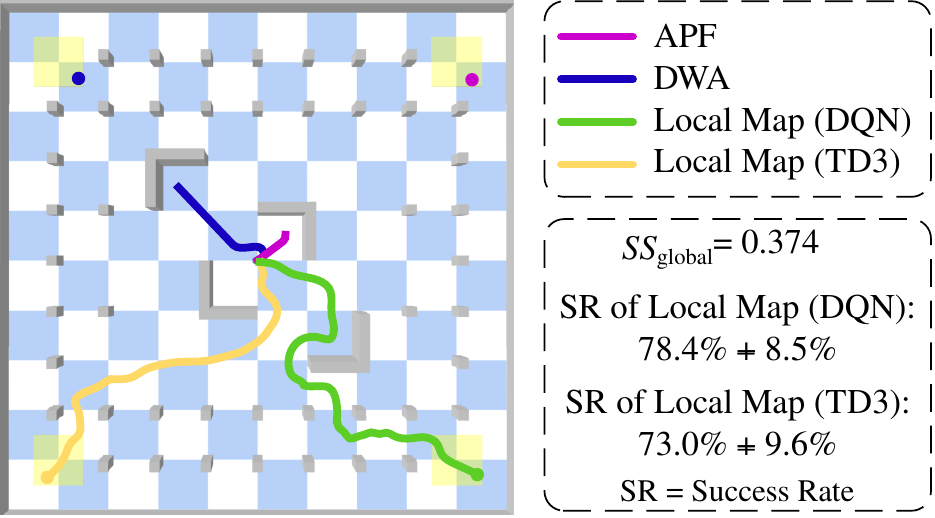}}
\caption{The sampled trajectories of the local map-based DRL navigation, APF\cite{apf} and DWA\cite{dwa}. Goal positions are randomly generated within the yellow area.}
\label{comp}
\end{figure}

A new simulation test scene is constructed to compare the local map-based DRL navigation with the traditional local planners APF\cite{apf} and DWA\cite{dwa}. We use the same agents trained in Section~\ref{com_vs_other} and the results are shown in Fig.\ref{comp}. APF and DWA fail to reach destinations due to the local optimum, while the local map reaches a success rate of over 70\% in the test scene with a low $SS_{\rm global}$ (0.374). When it is trained in a scene similar to the test scene, it is able to achieve better navigation performance. The results confirm the generalization of the local map-based DRL navigation in a new environment.

\section{Discussion}

\subsection{Comparision With Other Metrics}

\begin{figure}[tpb]
\centerline{\includegraphics[width=1.0\linewidth]{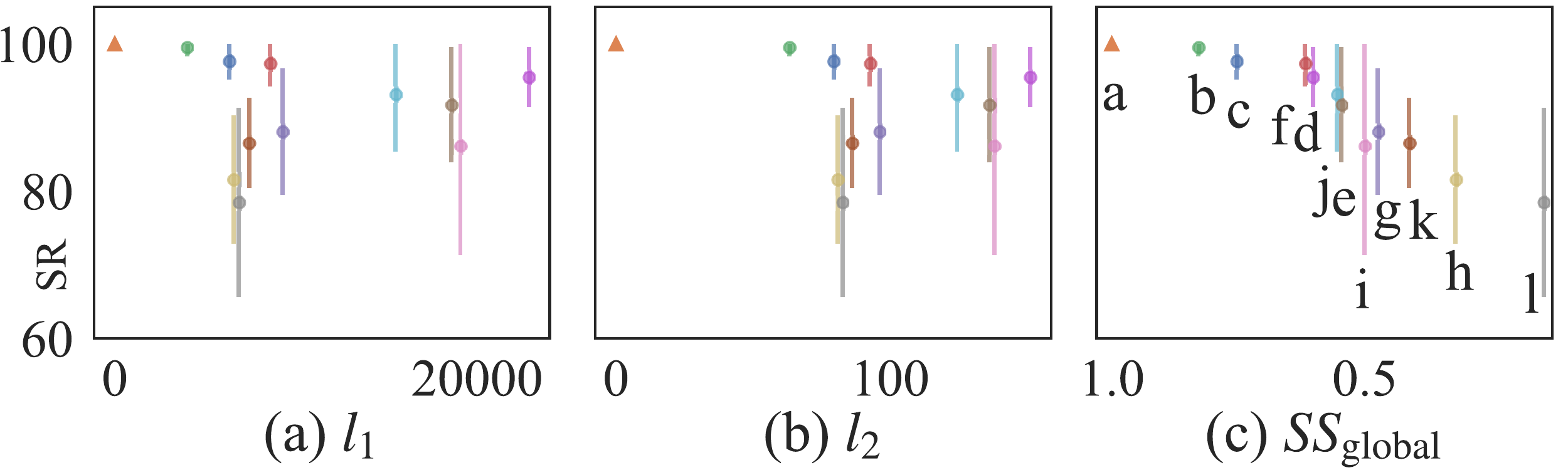}}
\caption{The statistical results of the relationship between the navigation success rates (mean and standard deviation) and different metrics (a) $l_1$ norm, (b) $l_2$ norm, and (c) $SS_{\rm global}$. The labels of the scenes are the same as in Fig.~\ref{scenarios}. The color representation of scenes is consistent with Fig.~\ref{simulaiton}(d). }
\label{metrics}
\end{figure}

To demonstrate the advantage of $SS_{\rm global}$ and $SS_{\rm local}$, the proposed metrics are compared to the $l_1$ and $l_2$ norms used in the discrepancy function of\cite{dr1}. The $l_1$ and $l_2$ norms require the same size of the global maps of scenes to be flattened into vectors with the same dimension, while the proposed transferability metric has no such limitation. We utilized the same data in Fig.~\ref{simulaiton} and the results of the relationship between navigation success rates and different metrics are presented in Fig.~\ref{metrics}. There exists no clear correlation between the navigation success rate and the $l_1$ and $l_2$ norms (Fig.~\ref{metrics}(a) and (b)). In comparison, as $SS_{\rm global}$ and $SS_{\rm local}$ decrease, the success rate decreases with a larger deviation (Fig.~\ref{metrics}(c) and Fig.~\ref{simulaiton}(d)). The results demonstrate the effectiveness of the proposed transferability metric.

\begin{figure}[t]
\centerline{\includegraphics[width=1.0\linewidth]{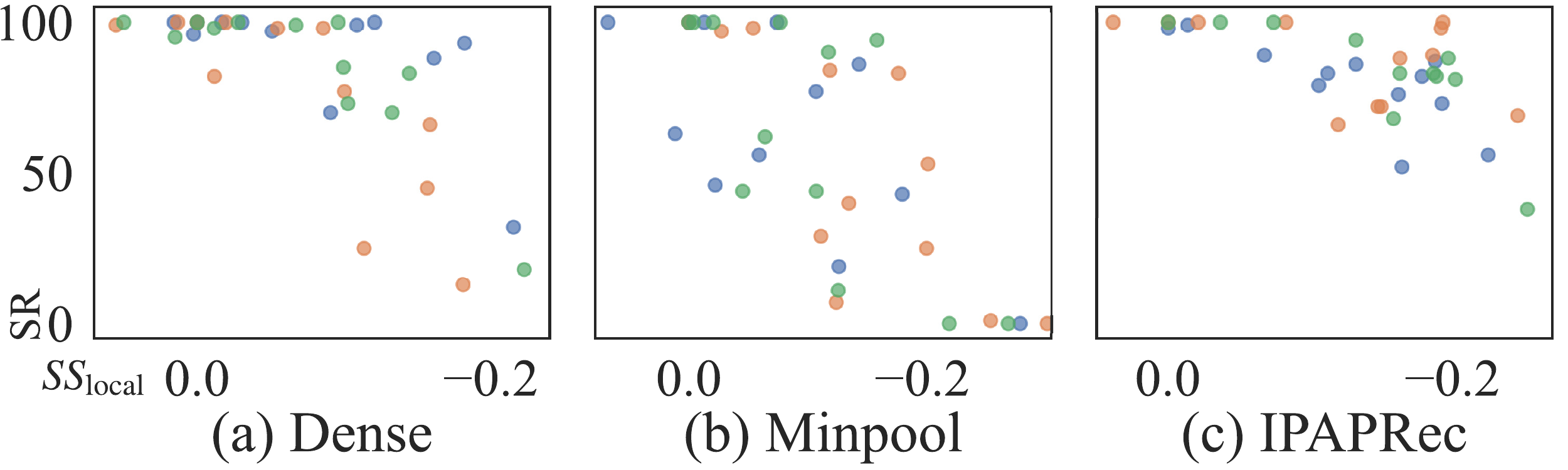}}
\caption{The navigation success rates (SR) of (a) Dense, (b) Minpool and (c) IPAPRec observations vs $SS_{\rm local}$ in test scenes shown in Fig.~\ref{scenarios}. Different colors represent different agents}
\label{gener}
\end{figure}

\subsection{Generalization of Local Scene Similarity}

To test the generalization of the local scene similarity, 3 independent agents in simulation using DQN for each observation including Dense, Minpool and IPAPRec with the same setting in Section~\ref{com_vs_other} are trained, since they use the same amount of LiDAR data as the local map. The success rate with respect to $SS_{\rm local}$ is shown in Fig.~\ref{gener}. With decreasing $SS_{\rm local}$, the success rates of three observations show a decreasing trend with a wider distribution. Although the local scene similarity is designed to quantify the transferability of the local map-based navigation algorithm, the results show a great potential in applying $SS_{\rm local}$ to other observations.

\section{Conclusion}

This paper proposed a novel scene similarity metric using the improved image template matching algorithm for quantifying the transferability of the DRL navigation algorithm exampled by the global and local performance measures. In addition, a DRL algorithm using the local map as the observation was designed for the navigation of mobile robots. A case study with a wheeled robot was designed and extensive experiments were conducted in a total of 26 simulated and real-world test scenes. The experimental results confirmed the robustness and transferability of the proposed local map-based navigation algorithm and showed the strong correlation between the designed scene similarity metric and the success rate of the DRL navigation algorithm when applied to new environments. To implement the global scene similarity in the real world, a mapping module, such as gmapping\cite{gmapping}, is required to generate the global map of the test scene.

In future work, we plan to refine the transferability metric, following the findings in the case study to comprehensively consider the scene similarity and the scene complexity. In addition, we plan to apply the proposed metric to other DRL-based navigation algorithms.

\vspace{-33pt}
\begin{IEEEbiography}[{\includegraphics[width=1in,height=1.25in,clip,keepaspectratio]{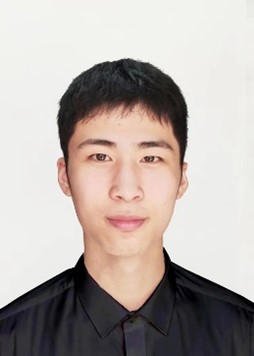}}]{Shiwei Lian}
received the Bachelor's degree in mechatronic engineering from Tongji University, Shanghai, China, in 2022. He is currently working toward the master’s degree in mechanical engineering with the College of Engineering, Peking University, China. 

His research interests include deep reinforcement learning and autonomous navigation.
\end{IEEEbiography}
\vspace{-33pt}

\begin{IEEEbiography}[{\includegraphics[width=1in,height=1.25in,clip,keepaspectratio]{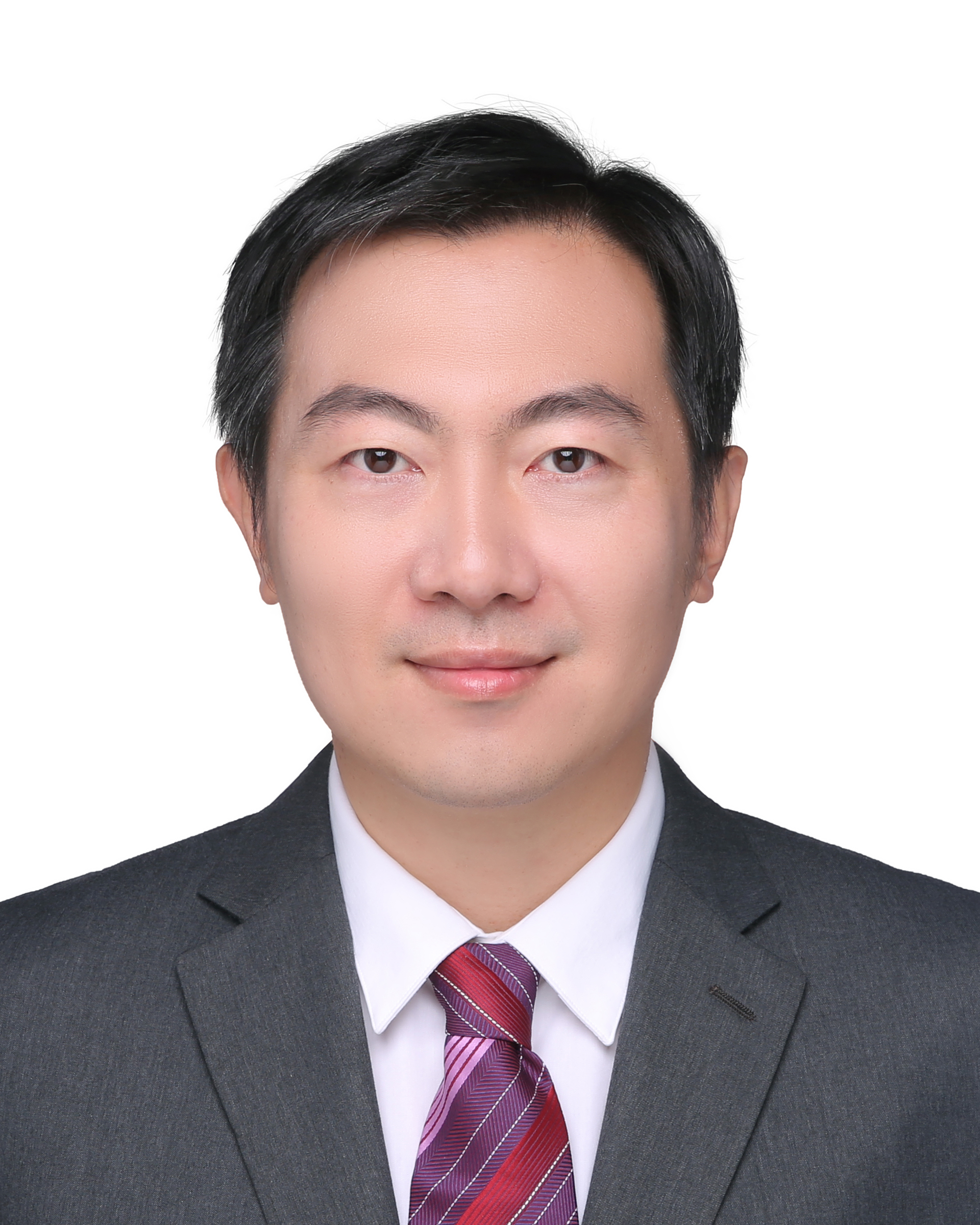}}]{Feitian Zhang}
(S’12–M’14) received the Bachelor’s and Master’s degrees in automatic control from Harbin Institute of Technology, Harbin, China, in 2007 and 2009, respectively, and the Ph.D. degree in electrical and computer engineering from Michigan State University, East Lansing, MI, USA, in 2014. He was a Postdoctoral Research Associate with the Department of Aerospace Engineering and Institute for Systems Research at University of Maryland, College Park, MD, USA from 2014 to 2016, and an Assistant Professor of Electrical and Computer Engineering with George Mason University, Fairfax, VA, USA from 2016 to 2021. He is currently an Associate Professor of Robotics Engineering with Peking University, Beijing, China. His research interests include mechatronics systems, robotics and controls, aerial vehicles and underwater vehicles.
\end{IEEEbiography}

\vfill

\end{document}